\pdfoutput=1
\documentclass[11pt]{article}

\usepackage[]{EACL2023}

\usepackage{times}
\usepackage{latexsym}
\usepackage{graphicx}
\usepackage{adjustbox}
\usepackage{amsmath}
\usepackage{amsfonts}
\usepackage{booktabs}
\usepackage{cprotect}

\usepackage{pgfplots}
\pgfplotsset{compat=1.16}

\usetikzlibrary{patterns}

\usepackage[T1]{fontenc}
\usepackage[utf8]{inputenc}

\usepackage{microtype}

\usepackage{inconsolata}

\title{An Extended Sequence Tagging Vocabulary for Grammatical Error Correction}

\author{
Stuart Mesham$^{\spadesuit}$ ~ Christopher Bryant$^{\diamondsuit}$ ~ Marek Rei$^{\heartsuit{}\diamondsuit}$ ~ Zheng Yuan$^{\clubsuit{}\diamondsuit}$
 \\
$^\spadesuit$Department of Computer Science, University of Cambridge \\
$^\diamondsuit$The ALTA Institute, Department of Computer Science, University of Cambridge\\ 
$^\heartsuit$Department of Computing, Imperial College London \\
$^\clubsuit$Department of Informatics, King's College London\\ 
{ \small \tt sm2613@cantab.ac.uk, christopher.bryant@cl.cam.ac.uk} \\
{ \small \tt marek.rei@imperial.ac.uk, zheng.yuan@kcl.ac.uk} \\
}

\begin{document}
\maketitle
\begin{abstract}
  We extend a current sequence-tagging approach to Grammatical Error Correction (GEC) by introducing specialised tags for spelling correction and morphological inflection using the SymSpell and LemmInflect algorithms. Our approach improves generalisation: the proposed new tagset allows a smaller number of tags to correct a larger range of errors. Our results show a performance improvement both overall and in the targeted error categories. We further show that ensembles trained with our new tagset outperform those trained with the baseline tagset on the public BEA benchmark.
\end{abstract}

\section{Introduction}
\label{sec:introduction}

Current approaches to Grammatical Error Correction (GEC) fall under two broad categories: sequence-to-sequence and sequence-tagging. The former treats GEC as a machine-translation problem, "translating" from error-containing to error-free language \citep{yuan-briscoe-2016-grammatical, schmaltz-etal-2017-adapting, junczys-dowmunt-etal-2018-approaching, grundkiewicz-etal-2019-neural, yuan2019neural, rothe-etal-2021-simple}. By contrast, sequence-tagging approaches tag each input word with an edit operation such that applying the operations produces the corrected output \citep{yannakoudakis2017neural, awasthi-etal-2019-parallel, omelianchuk-etal-2020-gector, tarnavskyi-etal-2022-ensembling}. The basic operations include keeping a word unchanged, deleting a word, and inserting new words \citep{awasthi-etal-2019-parallel, malmi-etal-2019-encode}.

One advantage of sequence-tagging over sequence-to-sequence approaches is computational efficiency: the former do not require expensive auto-regressive decoding,\footnote{\citet{malmi-etal-2019-encode} show that sequence-taggers can be orders of magnitude faster than comparable seq-to-seq models at inference time.} and currently achieve competitive performance using smaller models \citep{tarnavskyi-etal-2022-ensembling,rothe-etal-2021-simple}. However, current sequence-tagging approaches require manual linguistic efforts to curate language-specific edit tags~\citep{yuan-etal-2021-multi}. For example, \citet{awasthi-etal-2019-parallel} introduce rule-based morphological inflection tags, like replacing the "-ing" suffix with "-ion" (e.g. completing → completion). \citet{omelianchuk-etal-2020-gector} introduce a wider range of operations including verb-form and noun-number changes. For verb-form inflections, they use a dictionary to map between verb forms.\footnote{\url{https://github.com/gutfeeling/word_forms/blob/master/word_forms/en-verbs.txt}}

\begin{figure}[t]
    % \begin{adjustbox}{width=\textwidth}
    \includegraphics[width=\columnwidth]{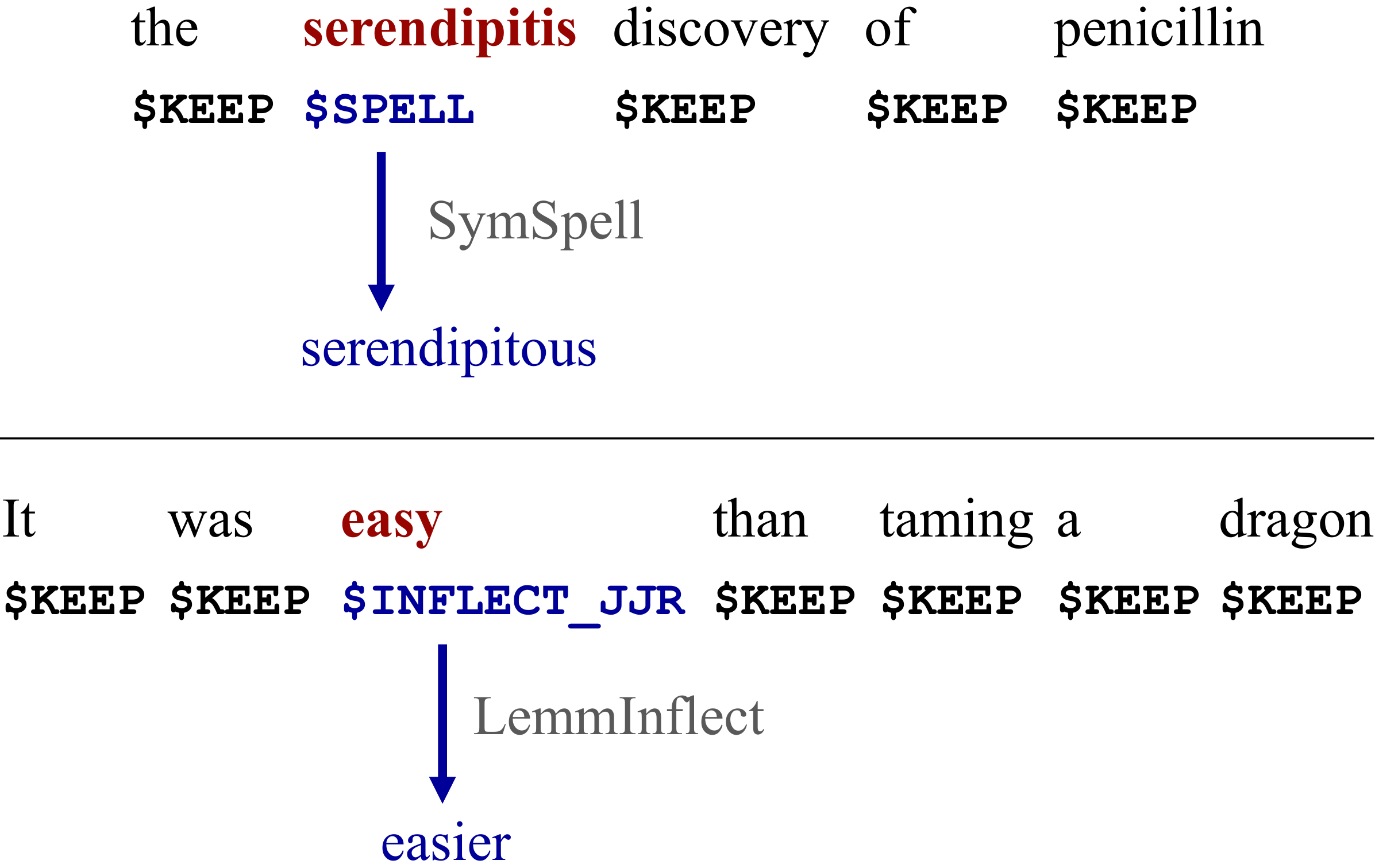}
    % \end{adjustbox}
    \cprotect\caption{\label{fig:diagram} Our model applied to two inputs. Beneath each word is the tagger's output. Arrows denote transformations by SymSpell and LemmInflect respectively.}
\end{figure}

In this paper, we focus on a sequence-tagging approach. We extend the approach of \citet{omelianchuk-etal-2020-gector} by introducing more general transformation tags (Figure~\ref{fig:diagram}). Specifically, we introduce:

\begin{itemize}
  \item A tag for correcting spelling errors.
  \item Inflection tags capable of a broader range of inflections than the tags introduced by \citet{omelianchuk-etal-2020-gector}.
\end{itemize}

These modifications allow a broader range of errors to be handled by a smaller number of transformation tags, which simplifies the sequence tagging problem, as well as improves the generalisation of the GEC system. Our results show that our modifications improve the system's performance on the BEA-2019 development and test sets. Our code and model weights are publicly available.\footnote{\url{https://github.com/StuartMesham/gector_experiment_public}}

\section{Methods}

We extend the system described by \citet{omelianchuk-etal-2020-gector} by adding new tags to the model's output vocabulary and modifying the inference and dataset preprocessing code to support our new tags. Our new tags perform spelling correction and morphological inflection and are described in Sections \ref{sec:spelling-tag} and \ref{sec:inflection-tags} below. We evaluate our tagset using the RoBERTa \citep{liu-2019-roberta}, DeBERTa \citep{he-2021-deberta}, DeBERTaV3 \citep{he-2021-debertav3}, ELECTRA \citep{clark-2020-electra} and XLNet \citep{yang-2019-xlnet} encoders, as well as an ensemble of three encoders (see Section~\ref{sec:ensembling}).

\subsection{Model and Training Procedure}

Our work builds on GECToR from \citet{omelianchuk-etal-2020-gector}, which follows the sequence tagging approach to GEC. We use the same sequence tagger architecture: a pre-trained transformer encoder with two separate "tagging" and "detection" heads. We also follow the same multi-phase training procedure using the synthetic PIE Corpus \citep{awasthi-etal-2019-parallel}, NUCLE \citep{dahlmeier-etal-2013-building}, FCE \citep{yannakoudakis-etal-2011-new}, Lang-8 \citep{mizumoto-etal-2011-mining, tajiri-etal-2012-tense} and W\&I + LOCNESS \citep{bryant-etal-2019-bea} English datasets.

\subsection{Baseline Tagset}
GECToR's tagset includes the basic edit tags, \verb|$KEEP|, \verb|$DELETE|, \verb|$REPLACE_{t}| and \verb|$APPEND_{t}|, which respectively leave the word unchanged, delete the word, replace the word with another word $t$, and append $t$ after the input word.

The tagset also contains a set of more complex grammatical transformation or ``g-transform'' tags. These include case, agreement (singular/plural), verb-form and merge/split transformations. For example, there is a tag to transform a verb into its past-tense equivalent. The verb-form transformations are performed using a dictionary. \citet[Table~9]{omelianchuk-etal-2020-gector} provide a full list of the transformations and their descriptions.

\subsection{Spelling Correction Tag}
\label{sec:spelling-tag}

GECToR corrects spelling errors using its vocabulary of \verb|$REPLACE_{t}| tags. This limits its ability to generalise to unseen or rare spelling errors for two reasons. The first is that GECToR can only correct misspellings of words which appear in its output vocabulary. The second is that for each word, there are many possible misspellings that the model must learn to associate with the corrected form.

To remedy this, we introduce a new \verb|$SPELL| tag for spelling correction. When this tag is predicted during inference, we use SymSpell\footnote{\url{https://github.com/wolfgarbe/SymSpell\#single-word-spelling-correction}} to produce the corrected version of the input word (see Section~\ref{sec:symspell} for details). We hypothesise that this improves generalisation because the sequence tagger need only detect spelling errors, and the corrections are performed by SymSpell. SymSpell can handle a variety of misspellings of each word and can correct words from a dictionary much larger than the output vocabulary of the sequence tagger.

% \usepackage{booktabs}
\begin{table*}[t]
    \centering
    \begin{adjustbox}{width=\textwidth,center}
    \begin{tabular}{@{}p{18em}|ccc|ccc@{}}
    \toprule
                                                                 & \multicolumn{3}{c|}{BEA-2019 dev}                                                  & \multicolumn{3}{c}{BEA-2019 test}                                                                \\
    Model                                                        & precision                 & recall                    & $F_{0.5}$                  & precision                 & recall                    & $F_{0.5} \text{ } (\bar{x} \pm \sigma)$  \\ \midrule
    $\text{DeBERTa}_{5K}^{(L)}$ basetags                         & 68.13                     & 38.12                     & 58.86                      & 77.89                     & 56.72                     & 72.47 ± 0.56                             \\
    $\text{DeBERTa}_{5K}^{(L)}$ \verb|$SPELL|                    & 68.37                     & \textbf{39.03}            & 59.40                      & 77.96                     & \textbf{57.67}            & 72.82 ± 0.49                             \\
    $\text{DeBERTa}_{5K}^{(L)}$ \verb|$INFLECT|                  & 68.73                     & 38.43                     & 59.33                      & 77.72                     & 57.23                     & 72.51 ± 0.93                             \\
    $\text{DeBERTa}_{5K}^{(L)}$ \verb|$SPELL| + \verb|$INFLECT|  & \textbf{69.75}            & 38.97                     & \textbf{60.20}             & \textbf{78.45}            & 57.44                     & \textbf{73.09 ± 0.72}                    \\ \midrule
    ensemble basetags                                            & 73.25                     & 37.17                     & 61.32                      & 83.47                     & 55.64                     & 75.87 ± 0.20                             \\
    ensemble \verb|$SPELL|                                       & 73.54                     & 37.76                     & 61.79                      & \textbf{83.72}            & \textbf{56.28}            & \textbf{76.26 ± 0.37}                    \\
    ensemble \verb|$INFLECT|                                     & 73.89                     & 37.35                     & 61.80                      & 83.71                     & 55.68                     & 76.06 ± 0.43                             \\
    ensemble \verb|$SPELL| + \verb|$INFLECT|                     & \textbf{74.19}            & \textbf{38.16}            & \textbf{62.39}             & 83.59                     & 56.23                     & 76.17 ± 0.38                             \\ \midrule
    $\text{DeBERTa}_{10K}^{(L)} \bigoplus \text{RoBERTa}_{10K}^{(L)} \bigoplus \text{XLNet}_{5K}^{(L)}$ \citep{tarnavskyi-etal-2022-ensembling} & 70.32                     & 34.62                     & 58.30                      & 84.44                     & 54.42                     & 76.05                     \\
    $\text{RoBERTa}_{5K}^{(L)}$ (KD)  \citep{tarnavskyi-etal-2022-ensembling}                      & -                         & -                         & -                          & 80.70                     & 53.39                     & 73.21                     \\
    T5 xxl \citep{rothe-etal-2021-simple}           & -                         & -                         & -                          & -                         & -                         & 75.88                     \\
    ESC \citep{qorib-etal-2022-frustratingly}       & \textbf{73.63}            & \textbf{40.12}            & \textbf{63.09}             & \textbf{86.65}            & \textbf{60.91}            & \textbf{79.90}            \\ \bottomrule
    \end{tabular}
    \end{adjustbox}
    \caption{A table showing BEA-2019 development and test set scores. The top section shows our models with varying tagsets using the $\text{DeBERTa}_{5K}^{(L)}$ encoder. The middle section shows the results for our ensemble models with varying tagsets. In the table, "ensemble" denotes the encoders $\text{DeBERTa}_{5K}^{(L)} \bigoplus \text{ELECTRA}_{5K}^{(L)} \bigoplus \text{RoBERTa}_{5K}^{(L)}$. Finally, the bottom section shows models from related work. The model labelled "(KD)" was trained using \citet{tarnavskyi-etal-2022-ensembling}'s knowledge distillation procedure. The results in the top and middle sections are averaged over 6 seeds, and the standard deviation, $\sigma$, of the test $F_{0.5}$ is shown.}
    \label{tab:combined-bea-results}
    \end{table*}

\subsection{Inflection Tags}
\label{sec:inflection-tags}

We introduce inflection tags of the form \verb|$INFLECT_{POS}| where POS denotes the Penn Treebank POS tag of the desired form of the input word. When an inflection tag is predicted at inference time, the input word is inflected to the target POS specified in the tag. The inflection is achieved using the software modules spaCy\footnote{\url{https://spacy.io}} and LemmInflect\footnote{\url{https://github.com/bjascob/LemmInflect}}. LemmInflect first attempts to use a dictionary for the inflection. If the input word is not in LemmInflect's dictionary, the inflection is performed using a rule-based approach (see Section~\ref{sec:lemminflect} for details).

Our inflection tags offer two main advantages over GECToR's dictionary-based verb transformations. The first is that they are not limited to verbs, but rather can be used for any inflected part of speech.\footnote{In English, the inflected parts of speech are adjectives, adverbs, nouns and verbs.} The second is that words which do not appear in LemmInflect's dictionary can still be handled using a rule-based approach (see Section~\ref{sec:lemminflect}). We also note that GECToR's singular/plural transformation tag only adds or removes an "-s" from the end of the input word, making it unable to handle less trivial cases such as inflecting "activity" to its plural "activities". By contrast, our system applies the full dictionary and rule-based procedure to singular/plural transformations. In summary, our inflection tags handle a broader range of transformations than GECToR's transformation tags. We hypothesise that this improves generalisation.

\subsection{Preprocessing}
\label{sec:preprocessing}

To incorporate our \verb|$SPELL| tag into the training data, we take data preprocessed with \citet{omelianchuk-etal-2020-gector}'s code, and for each instance of a \verb|$REPLACE_{t}| tag, we apply SymSpell to the input word. If SymSpell produces the correct output, $t$, we change the \verb|$REPLACE_{t}| tag to a \verb|$SPELL| tag. Otherwise, we leave the \verb|$REPLACE_{t}| tag unchanged.

For the inflection tags, we first modify \citet{omelianchuk-etal-2020-gector}'s preprocessing code by removing existing tags which perform inflections.\footnote{We remove tags g-8 to g-29 \citep[Table~9]{omelianchuk-etal-2020-gector}.} Then, similar to our process for the \verb|$SPELL| tag, for each instance of a \verb|$REPLACE_{t}| tag, we attempt to inflect the input word to obtain the target word $t$ and, if successful, change the tag to an \verb|$INFLECT_{POS}| tag. Otherwise, we leave the tag unchanged. For details about this process, we refer the reader to the relevant script in our repository.\cprotect\footnote{See the \verb|lemminflect_preprocess.py| script in the \verb|utils| directory of our repository.}

\subsection{Ensembling}
\label{sec:ensembling}

To create ensemble models, we use the span-based voting procedure of \citet{tarnavskyi-etal-2022-ensembling}. Their system takes the corrected output of each model, compares it with the input text, and extracts edit spans of the same type (insert, delete, or replace). In an ensemble of $k$ models, spans predicted by at least $k-1$ models are included in the output of the ensemble.

Our particular combination of encoders was chosen on the BEA-2019 development set by searching over all possible combinations of three models from the set of individual models we trained with the \verb|$SPELL| + \verb|$INFLECT| tagset.

\section{Results}

\begin{figure*}[t]
    \centering
    \begin{tikzpicture}
        \begin{axis}[
        ybar, % draw a bar chart (not a line graph)
        symbolic x coords={deberta-large-10k,deberta-large-5k,deberta-v3-large-10k,deberta-v3-large-5k,electra-large-10k,electra-large-5k,roberta-large-10k,roberta-large-5k,xlnet-large-10k,xlnet-large-5k},
        xtick=data,
        ylabel={$F_{0.5}$},
        bar width=0.3cm,
        x=1.3cm, % "unit distance" on x axis
        xticklabels={$\text{DeBERTa}_{10K}^{(L)}$,$\text{DeBERTa}_{5K}^{(L)}$,$\text{DeBERTaV3}_{10K}^{(L)}$,$\text{DeBERTaV3}_{5K}^{(L)}$,$\text{ELECTRA}_{10K}^{(L)}$,$\text{ELECTRA}_{5K}^{(L)}$,$\text{RoBERTa}_{10K}^{(L)}$,$\text{RoBERTa}_{5K}^{(L)}$,$\text{XLNet}_{10K}^{(L)}$,$\text{XLNet}_{5K}^{(L)}$},
        x tick label style={rotate=-45,anchor=west},
        legend pos=north east
        ]
        \addplot [color=blue, pattern color=blue, pattern=dots, error bars/.cd, y dir=both,y explicit, x dir=both,x explicit, error bar style=black] coordinates {(deberta-large-10k, 72.65607401102228) +-=(0, 0.8791480660644746) (deberta-large-5k, 70.45457455209834) +-=(0, 1.8655802973216709) (deberta-v3-large-10k, 71.14544028798024) +-=(0, 0.6499477954708237) (deberta-v3-large-5k, 68.90033215823783) +-=(0, 0.7006697855985702) (electra-large-10k, 66.10492327949588) +-=(0, 0.10251129443536913) (electra-large-5k, 66.45708097902559) +-=(0, 0.7820239118080341) (roberta-large-10k, 72.81538400834818) +-=(0, 0.892317492785464) (roberta-large-5k, 70.420406276317) +-=(0, 1.787003292524701) (xlnet-large-10k, 69.83617716756078) +-=(0, 1.7252132382120802) (xlnet-large-5k, 66.75290458351368) +-=(0, 1.36643875311878) };
        \addplot [color=orange, pattern color=orange, pattern=north west lines, error bars/.cd, y dir=both,y explicit, x dir=both,x explicit, error bar style=black] coordinates {(deberta-large-10k, 76.4890321293221) +-=(0, 1.6284385317962162) (deberta-large-5k, 75.80337666181885) +-=(0, 1.1622575020957664) (deberta-v3-large-10k, 73.87396217488329) +-=(0, 0.4809017472925627) (deberta-v3-large-5k, 73.11440507798358) +-=(0, 1.413420137212869) (electra-large-10k, 73.64598456088093) +-=(0, 0.962645532159955) (electra-large-5k, 73.21216295941643) +-=(0, 0.5332372814112) (roberta-large-10k, 76.07360241968344) +-=(0, 1.2003833470216851) (roberta-large-5k, 76.43834142850388) +-=(0, 0.18301272464270765) (xlnet-large-10k, 73.70990710570314) +-=(0, 0.9976494230421358) (xlnet-large-5k, 73.05852898791956) +-=(0, 1.1350258449065072) };
        \legend{basetags, \texttt{\$SPELL}}
        \end{axis}
    \end{tikzpicture}
    \cprotect\caption{\label{fig:spell-target-error-scores} A bar graph showing the BEA-2019 development set $F_{0.5}$ scores for the "spelling" error category for different encoders, tagsets and vocabulary sizes. Specifically, the \verb|$SPELL| and basetags tagsets and vocabulary sizes of 5k and 10k. Each bar represents the mean score over three training runs with different seeds. The error bars show the standard deviations of the scores.}
\end{figure*}
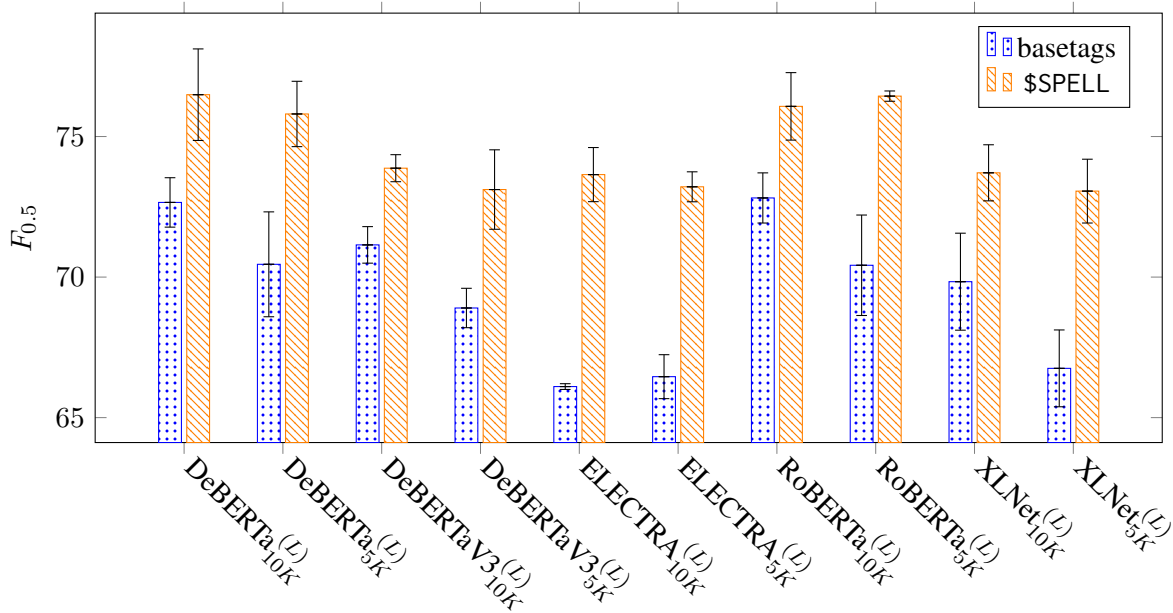

We report the span-based precision, recall and $F_{0.5}$ scores on the BEA-2019 development and test sets \citep{bryant-etal-2019-bea} using the ERRANT scorer \citep{bryant-etal-2017-automatic}.\footnote{\url{https://github.com/chrisjbryant/errant}} The term "basetags" indicates the tagset proposed by \citet{omelianchuk-etal-2020-gector}, and \verb|$SPELL| and \verb|$INFLECT| denote our proposed tagsets containing the spelling and inflection tags respectively. \verb|$SPELL| + \verb|$INFLECT| denotes tagsets containing both the spelling and inflection tags. We adopt the model and tagset size notation of \citet{tarnavskyi-etal-2022-ensembling} which, for example, denotes a DeBERTa-large model using a 5k vocabulary size as $\text{DeBERTa}_{5K}^{(L)}$.

Table~\ref{tab:combined-bea-results} shows the scores of our models on the BEA-2019 development and test sets. Of the three encoders chosen for our ensemble, $\text{DeBERTa}_{5K}^{(L)}$ had the highest mean development set score when using \verb|$SPELL| + \verb|$INFLECT| tagset, and is thus shown in Table~\ref{tab:combined-bea-results}.\footnote{See Section~\ref{sec:additional-single-encoder-results} for the results of the other encoders, and Section~\ref{sec:conll-2014-results} for CoNLL-2014 results.}

For the $\text{DeBERTa}_{5K}^{(L)}$ encoder, on both the development and test sets, the \verb|$SPELL| and \verb|$INFLECT| tagsets provide an improvement over the basetags tagset, and the \verb|$SPELL| + \verb|$INFLECT| tagset provides a larger improvement. Similarly, for the ensemble models, on the development set, the \verb|$SPELL| and \verb|$INFLECT| tagsets show an improvement over the basetags tagset, and the \verb|$SPELL| + \verb|$INFLECT| tagset obtains the highest score. However, on the test set, the \verb|$SPELL| tagset scores the highest.

\subsection{Target Error Categories}
\label{sec:target-error-categories}

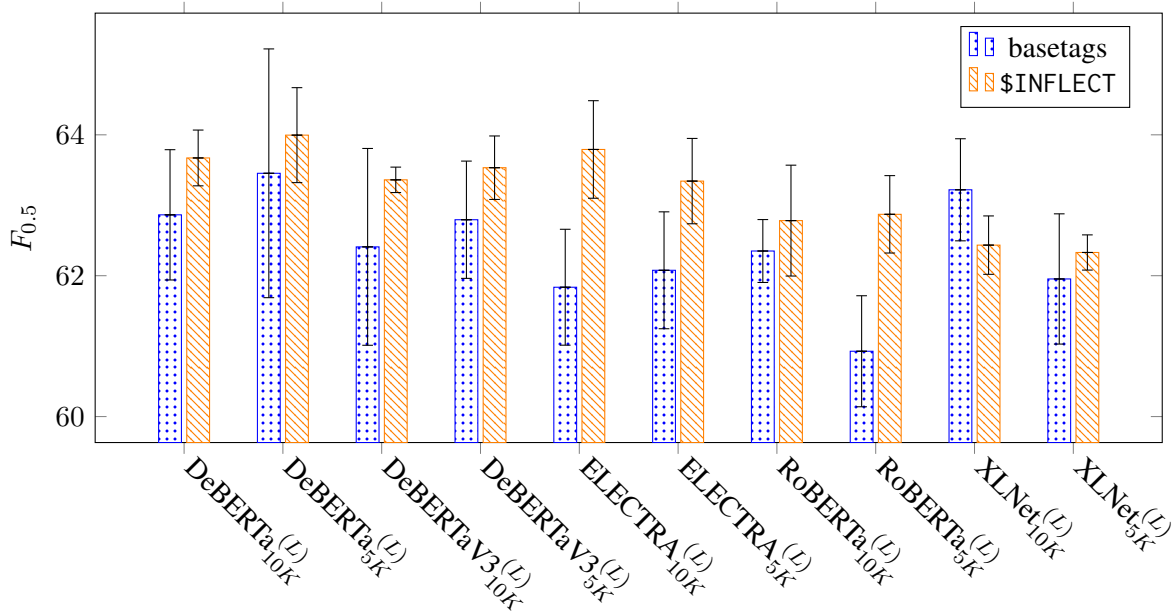
\begin{figure*}[t]
    \centering
    \begin{tikzpicture}
        \begin{axis}[
        ybar, % draw a bar chart (not a line graph)
        symbolic x coords={deberta-large-10k,deberta-large-5k,deberta-v3-large-10k,deberta-v3-large-5k,electra-large-10k,electra-large-5k,roberta-large-10k,roberta-large-5k,xlnet-large-10k,xlnet-large-5k},
        xtick=data,
        ylabel={$F_{0.5}$},
        bar width=0.3cm,
        x=1.3cm, % "unit distance" on x axis
        xticklabels={$\text{DeBERTa}_{10K}^{(L)}$,$\text{DeBERTa}_{5K}^{(L)}$,$\text{DeBERTaV3}_{10K}^{(L)}$,$\text{DeBERTaV3}_{5K}^{(L)}$,$\text{ELECTRA}_{10K}^{(L)}$,$\text{ELECTRA}_{5K}^{(L)}$,$\text{RoBERTa}_{10K}^{(L)}$,$\text{RoBERTa}_{5K}^{(L)}$,$\text{XLNet}_{10K}^{(L)}$,$\text{XLNet}_{5K}^{(L)}$},
        x tick label style={rotate=-45,anchor=west},
        legend pos=north east
        ]
        \addplot [color=blue, pattern color=blue, pattern=dots, error bars/.cd, y dir=both,y explicit, x dir=both,x explicit, error bar style=black] coordinates {(deberta-large-10k, 62.864252881699336) +-=(0, 0.9247864020828789) (deberta-large-5k, 63.45396315255508) +-=(0, 1.7658127650718305) (deberta-v3-large-10k, 62.409465121689855) +-=(0, 1.3970719140834424) (deberta-v3-large-5k, 62.79472593045026) +-=(0, 0.8324616521652561) (electra-large-10k, 61.83646624786602) +-=(0, 0.8227708777349061) (electra-large-5k, 62.07700829839456) +-=(0, 0.8298090921386276) (roberta-large-10k, 62.35060143620533) +-=(0, 0.44645270200824694) (roberta-large-5k, 60.92758119434367) +-=(0, 0.788043524686449) (xlnet-large-10k, 63.21989316637613) +-=(0, 0.7245924023612952) (xlnet-large-5k, 61.95399312087998) +-=(0, 0.9244982922353983) };
        \addplot [color=orange, pattern color=orange, pattern=north west lines, error bars/.cd, y dir=both,y explicit, x dir=both,x explicit, error bar style=black] coordinates {(deberta-large-10k, 63.6719100630715) +-=(0, 0.3958624797679815) (deberta-large-5k, 63.995722054695705) +-=(0, 0.6740602040475918) (deberta-v3-large-10k, 63.361149898339754) +-=(0, 0.18053088350080315) (deberta-v3-large-5k, 63.533407301395776) +-=(0, 0.45053524307835063) (electra-large-10k, 63.79229866068596) +-=(0, 0.6924183735507249) (electra-large-5k, 63.34312775411013) +-=(0, 0.6058221924081562) (roberta-large-10k, 62.78238909151501) +-=(0, 0.7868644035744631) (roberta-large-5k, 62.871561933203225) +-=(0, 0.5493868698744975) (xlnet-large-10k, 62.434264234570435) +-=(0, 0.41450460458773974) (xlnet-large-5k, 62.32975710300164) +-=(0, 0.249937403266264) };
        \legend{basetags, \texttt{\$INFLECT}}
        \end{axis}
    \end{tikzpicture}
    \cprotect\caption{\label{fig:lemon-target-error-scores} A bar graph showing the BEA-2019 development set $F_{0.5}$ scores for inflection-related errors for different encoders, tagsets and vocabulary sizes. Specifically, the \verb|$INFLECT| and basetags tagsets and vocabulary sizes of 5k and 10k. Each bar represents the mean score over three training runs with different seeds. The error bars show the standard deviations of the scores.}
\end{figure*}

Figures \ref{fig:spell-target-error-scores} and \ref{fig:lemon-target-error-scores} show BEA-2019 development set scores in the ERRANT error categories \citep[Table~2]{bryant-etal-2017-automatic} targeted by the \verb|$SPELL| and \verb|$INFLECT| tagsets respectively. The former targets only the "spelling" error category, and the latter targets categories related to inflection.\footnote{Specifically, the ADJ:FORM, MORPH, NOUN:INFL, NOUN:NUM, VERB:FORM, VERB:INFL, VERB:SVA and VERB:TENSE categories.} In Figure~\ref{fig:spell-target-error-scores} we observe substantial performance improvements in the spelling category for all models. Figure~\ref{fig:lemon-target-error-scores} shows a smaller improvement in the target error categories of the \verb|$INFLECT| tagset for all models except $\text{XLNet}_{10K}^{(L)}$.

\section{Discussion}

In general, the \verb|$SPELL| and \verb|$INFLECT| tagsets both improve performance over the baseline tagset. The results of Section~\ref{sec:target-error-categories} show that the tagsets improve performance in their respective targeted error categories. This indicates that our modifications were successful.

In the results showing all error categories (Table~\ref{tab:combined-bea-results}), the inclusion of many non-targeted categories reduces the weighting of targeted categories, resulting in smaller apparent differences between models. For the ensemble models, the \verb|$SPELL| tagset obtains a higher test score than the \verb|$SPELL| + \verb|$INFLECT| tagset. This is contrary to our expectation that the combination of our modifications should provide a cumulative improvement. It is also unexpected that the ranking of the ensemble models on the development and test sets differs.

Differences in error-type frequencies in the development and test sets do not provide an explanation, since the frequency of spelling errors is lower in the test set than in the development set, and the frequencies of the error types which the \verb|$INFLECT| tagset most impacts\footnote{Specifically the NOUN:NUM, VERB:FORM and VERB:SVA error types. See Section~\ref{sec:detailed-inflection-error-analysis} for details.} are higher in the test set than in the development set \citep[Table 4]{bryant-etal-2019-bea}. We therefore hypothesise that this unexpected pattern is an artefact of the variation between different random seeds.

\section{Conclusions}

We have motivated and described new tags for spelling correction and morphological inflection. These tags are capable of correcting a broader range of errors than previous tags, thereby improving generalisation. Our results show that the new tags improve performance both in the targeted error categories and overall for both single-encoder models and ensembles.

Our findings ultimately show there is great scope for improving GEC sequence-labelling model performance by introducing tags capable of correcting more general and possibly complex classes of errors.

Finally, we believe our results are of immediate value to practitioners building GEC applications since they offer improved performance without the use of seq-to-seq models which can require orders of magnitude more computation at inference time.

\section{Future Work}

We used SymSpell in its context-free configuration when correcting spelling errors. We chose this method because of its speed and simplicity, however, better performance could likely be obtained by switching to a context-sensitive spelling correction algorithm.

Although our experiments demonstrate a performance improvement over the results of \citet{tarnavskyi-etal-2022-ensembling}, other recent work has demonstrated further performance improvements \citep{lai-etal-2022-type, qorib-etal-2022-frustratingly}. Our contribution is orthogonal to these, and so future work could investigate whether using our tagset for the sequence tagger used by \citet{lai-etal-2022-type} or using our models in the ensemble described by \citet{qorib-etal-2022-frustratingly} would yield further improvements.

\section*{Limitations}
The results obtained have high variance with respect to the random seed used (see Appendix Figures \ref{fig:ensemble-dev-single} and \ref{fig:ensemble-test-single}). Due to compute limitations, we were unable to run more seeds to better observe the distributions of development and test scores. 

The generalised tags we experimented with are also somewhat language specific, as, for example, the \verb|$INFLECT| tagset will not be beneficial to a language with little or no morphology.

\section*{Ethics Statement}

This work is conducted in accordance with the ACM Code of Ethics.\footnote{\url{https://www.acm.org/code-of-ethics}} In this section we comment on the topics of privacy, safety and accessibility, as we believe they are particularly relevant to the development and use of our system.

\subsection*{Privacy}

Since machine learning systems can reveal sensitive information about their training data, it is important to consider privacy concerns relating to the development and use of such systems. The training data for our system originates from two primary sources: publicly available text and essays collected from examinations and online error correction services. The PIE Corpus is derived from publicly available texts \citep{awasthi-etal-2019-parallel}. The Lang-8 and Write \& Improve essays are collected in accordance with the services' respective privacy policies. The FCE dataset is anonymised before use \citep{yannakoudakis-etal-2011-new}. Privacy-related information is not documented for the NUCLE and LOCNESS datasets.

\subsection*{Safety}

Automated GEC systems have the potential to change the meaning of the input text. Therefore, the systems described in this work should be applied with caution. In scenarios where miscommunication is dangerous, the system should only be used as an aid for the manual correction of text, rather than a fully automated system.

\subsection*{Accessibility}

The development of our system required compute-intensive model training and data preprocessing.\footnote{See Section~\ref{sec:model-size-compute} for details.} This cost may be prohibitive for some research groups or potential users. We make our trained models, hyperparameters and source code publicly available to alleviate this issue and increase the accessibility of our developments.

\section*{Acknowledgements}

This work was primarily funded by the Skye Foundation and Cambridge Trust.

\bibliography{anthology,custom}
\bibliographystyle{acl_natbib}

\appendix

\section{Appendix}
\label{sec:appendix}

\subsection{SymSpell}
\label{sec:symspell}

SymSpell is an open-source spelling correction system. It is initialised with a dictionary of correct words and their frequency in some sample of English text. Given a misspelt input word, the system searches its dictionary for the word with the minimum Damerau-Levenshtein distance \citep{damerau-1964-technique} from the input, breaking ties using the word frequencies. A parameter $n$ limits the maximum number of edits allowed. If the dictionary contains no words within $n$ Damerau-Levenshtein edits of the input, the system reports that the input could not be corrected.

We initialise SymSpell using $n=2$ and use the dictionary of approximately 83k English words included with SymSpell.\footnote{\url{https://github.com/wolfgarbe/SymSpell/blob/master/SymSpell/frequency_dictionary_en_82_765.txt}} The dictionary is derived from the Spell Checker Oriented Word Lists\footnote{\url{http://wordlist.aspell.net}} database and contains both British and American spelling variants. Word frequencies are obtained from the Google Books n-gram dataset.\footnote{\url{https://storage.googleapis.com/books/ngrams/books/datasetsv2.html}}

\subsection{LemmInflect}
\label{sec:lemminflect}

LemmInflect is a software module which performs lemmatisation and inflection on English words. For example, we may want to inflect the singular present tense verb ``runs'' to its past tense form ``ran''. We can do this by first computing the lemma of ``runs'' using \verb|getLemma('runs', upos='VERB')|, and then inflecting it to its past tense form using \verb|getInflection(lemma, tag='VBD')|, where \verb|lemma| is the output of the previous step.\footnote{The ``upos'' and ``tag'' arguments are the Universal POS tag \citep{nivre-etal-2020-universal} of the input word and the Penn Treebank POS tag \citep{marcus-etal-1993-building} of the desired output respectively.} LemmInflect's functions first attempt to use dictionaries to map between word forms. If the input does not appear in its dictionary, LemmInflect uses a classification model to determine which of a pre-defined set of morphing rules to apply (e.g. adding ``-ed'' to the input).

When an \verb|$INFLECT_{POS}| tag is predicted by our sequence tagger, the inflection is performed by first tagging the input sentence with Universal POS (UPOS) tags using spaCy, then computing the lemma of the input word with LemmInflect's \verb|getLemma| function. Finally, the lemma of the input word is inflected to the target POS using the \verb|getInflection| function.

\subsection{Training Details}

We use a batch size of 256 in stages 1 and 2, and 128 in stage 3. During training, the model is evaluated on the development set every 10k steps in stage one, and every epoch in stages two and three. Training is stopped when the development set accuracy does not improve for three consecutive evaluations or a maximum number of training steps or epochs have been completed. The accuracy is computed as the combined tag-level accuracy of the detection and tagging heads. We use a maximum of 200k steps for stage one, and a maximum of 15 epochs for stages two and three. In our experiments, stages two and three never reach this maximum.

We use the cross entropy loss function\footnote{\url{https://pytorch.org/docs/stable/generated/torch.nn.CrossEntropyLoss.html}} and the Adam optimiser \citep{kingma-2014-adam} with the default parameters ($\beta_1=0.9$, $\beta_2=0.999$, $\epsilon=10^{-8}$).\footnote{We use the PyTorch implementation of the AdamW optimiser \citep{loshchilov-2018-decoupled} with the \textit{weight decay} parameter set to zero, making it equivalent to the Adam optimiser.} We follow the learning rate schedule of \citep{omelianchuk-etal-2020-gector}. Specifically, we perform the first 20k steps and first 2 epochs of training stages one and two respectively with a learning rate of $10^{-3}$ and the encoder weights frozen.\footnote{During this initial phase, only the weights of the prediction heads are updated.} After these respective points in training are reached, the encoder weights are unfrozen and the learning rate is decreased to $10^{-5}$. In stage three, the encoder weights are never frozen and we only use a learning rate of $10^{-5}$.

Once a model has been trained, we perform a grid search on the BEA-2019 development set over the possible values of the \textit{confidence bias} and \textit{minimum error probability} parameters \citep{omelianchuk-etal-2020-gector}. We later refer to these as the "inference tweak" parameters. For both parameters, we test values ranging from 0.0 to 0.9 inclusive, in increments of 0.02, resulting in a total of 2116 ($46 \times 46$) development set evaluations of the model. We have included, in our public repository, the BEA-2019 development set scores for all of the parameter combinations tested, as well as the chosen parameters for each of the models.

\subsection{Dataset Sizes and Splits}

We use the same datasets for each training stage as \citet{omelianchuk-etal-2020-gector}. We refer readers to Table~1 of their paper for statistics on each dataset's size and error frequencies. For stages 1 and 2, we combine the relevant datasets as described in their repository.\footnote{\url{https://github.com/grammarly/gector/blob/master/docs/training_parameters.md}} We generate a random split of each dataset into training and development sets, which contain 98\% and 2\% of the data respectively.\footnote{The 98/2 training/development split was used by \citet{omelianchuk-etal-2020-gector}. This is documented in the main README file in their repository.} For stage 3, we use the pre-defined training, development and test sets of the W\&I + LOCNESS dataset \citep{bryant-etal-2019-bea}.

\subsection{Model Size and Compute Requirements}
\label{sec:model-size-compute}

We use the standard "large" configuration of each of our encoders. The number of parameters in each encoder is shown in Table~\ref{tab:model-sizes}.

\begin{table}[t]
\centering
\begin{tabular}{@{}lc@{}}
\toprule
Encoder         & Parameters \\ \midrule
DeBERTa-large   & 405M       \\
DeBERTaV3-large & 435M       \\
ELECTRA-large   & 335M       \\
RoBERTa-large   & 355M       \\
XLNet-large     & 360M       \\ \bottomrule
\end{tabular}
\caption{A table showing the number of parameters in each of the encoders we use. Note that these numbers do not include the weights of the detection and tagging heads which vary based on the vocabulary size used.}
\label{tab:model-sizes}
\end{table}

Training took 15-20 hours per model with four NVIDIA A100 GPUs connected via NVLink, each with 80 GB of VRAM, using the HuggingFace PyTorch DistributedDataParallel trainer implementation. Our grid search over the inference tweak hyperparameters took 8-13 hours on one A100.

We did not perform detailed inference time experiments. For inference jobs that were run on an NVIDIA A100 GPU using a batch size of 128, inference over the BEA-2019 development set took approximately 10s with the basetags and \verb|$SPELL| models and approximately 20s with the \verb|$INFLECT| and \verb|$SPELL| + \verb|$INFLECT| models. We note that our implementation was not optimised for inference speed. It processes \verb|$INFLECT| tags sequentially on a single CPU thread, whereas an optimised implementation could parallelise this processing within a batch of sentences. 

\begin{figure}[t]
    \includegraphics[width=\linewidth]{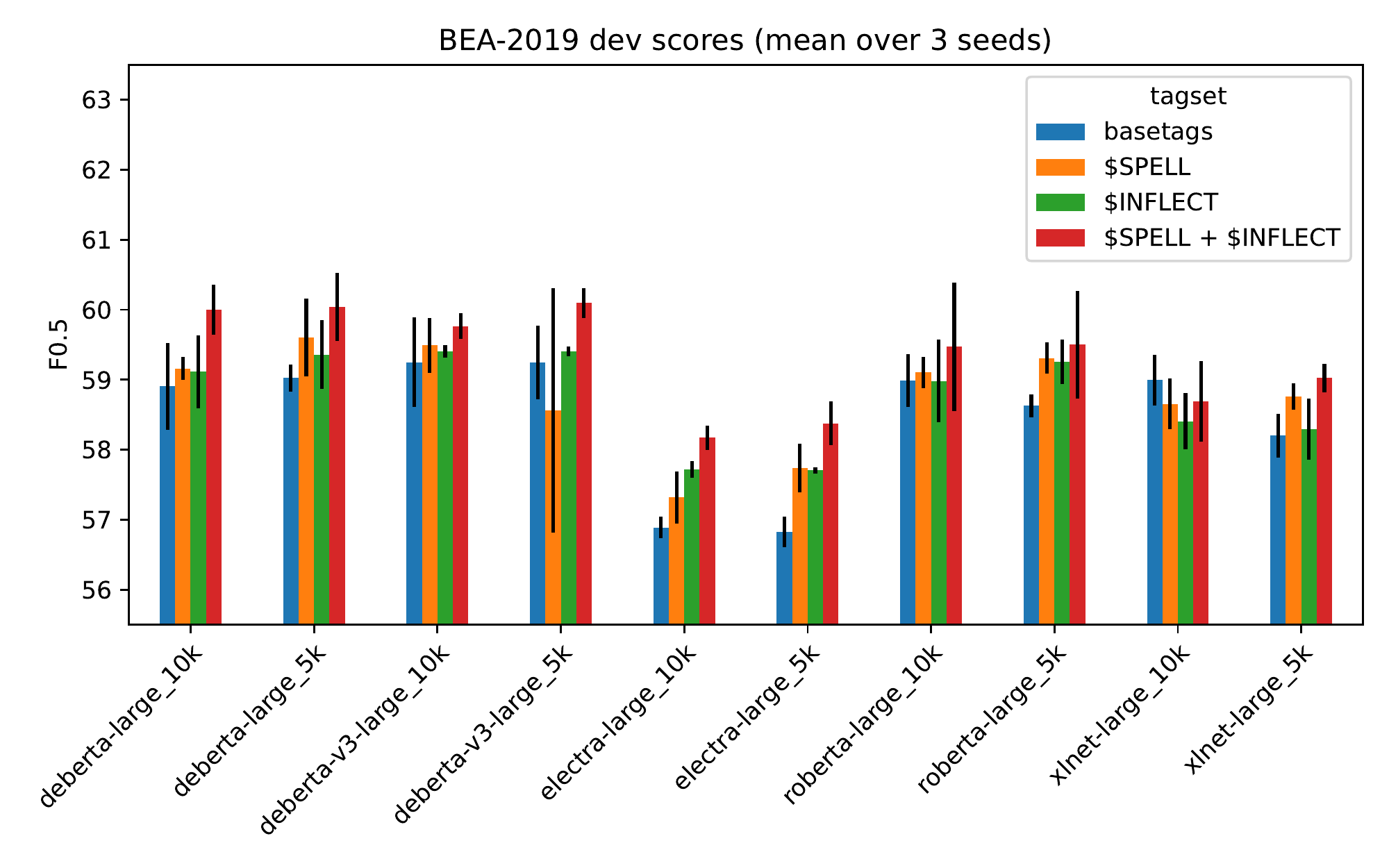}
    
    \caption{\label{fig:single-model-dev-mean} A bar graph showing the BEA-2019 development set $F_{0.5}$ scores of single models using different tagsets. Each bar represents the mean score over three training runs with different seeds. The black lines are error bars showing the standard deviations.}
\end{figure}

This paper reports results from 156 models\footnote{Figures \ref{fig:spell-target-error-scores}-\ref{fig:single-model-dev-mean} show the results from 120 models (5 encoders $\times$ 2 vocabulary sizes $\times$ 4 tagsets $\times$ 3 seeds) and Tables \ref{tab:combined-bea-results} and \ref{tab:combined-conll-results} required a further 36 models to be trained (3 encoders $\times$ 4 tagsets $\times$ 3 seeds).} which took approximately 12.5k GPU hours to train and tune. Before this, we used approximately 2k GPU hours for development and preliminary experiments with smaller models. Therefore in total, approximately 14.5k GPU hours were used in creating this paper.

The training data preprocessing for our new inflection tags is CPU-intensive because, for every sentence, both the input and approximated gold output need to be POS-tagged with spaCy, and LemmInflect needs to be applied to every \verb|$REPLACE_{t}| tag. In our experiments, preprocessing the datasets for all three training stages took approximately 35 minutes on a dual-socket 76-core Intel(R) Xeon(R) Platinum 8368Q CPU @ 2.60GHz. This process was run for both the \verb|$INFLECT| and \verb|$SPELL| + \verb|$INFLECT| tagsets.

\begin{figure}[t]
    \includegraphics[width=\linewidth]{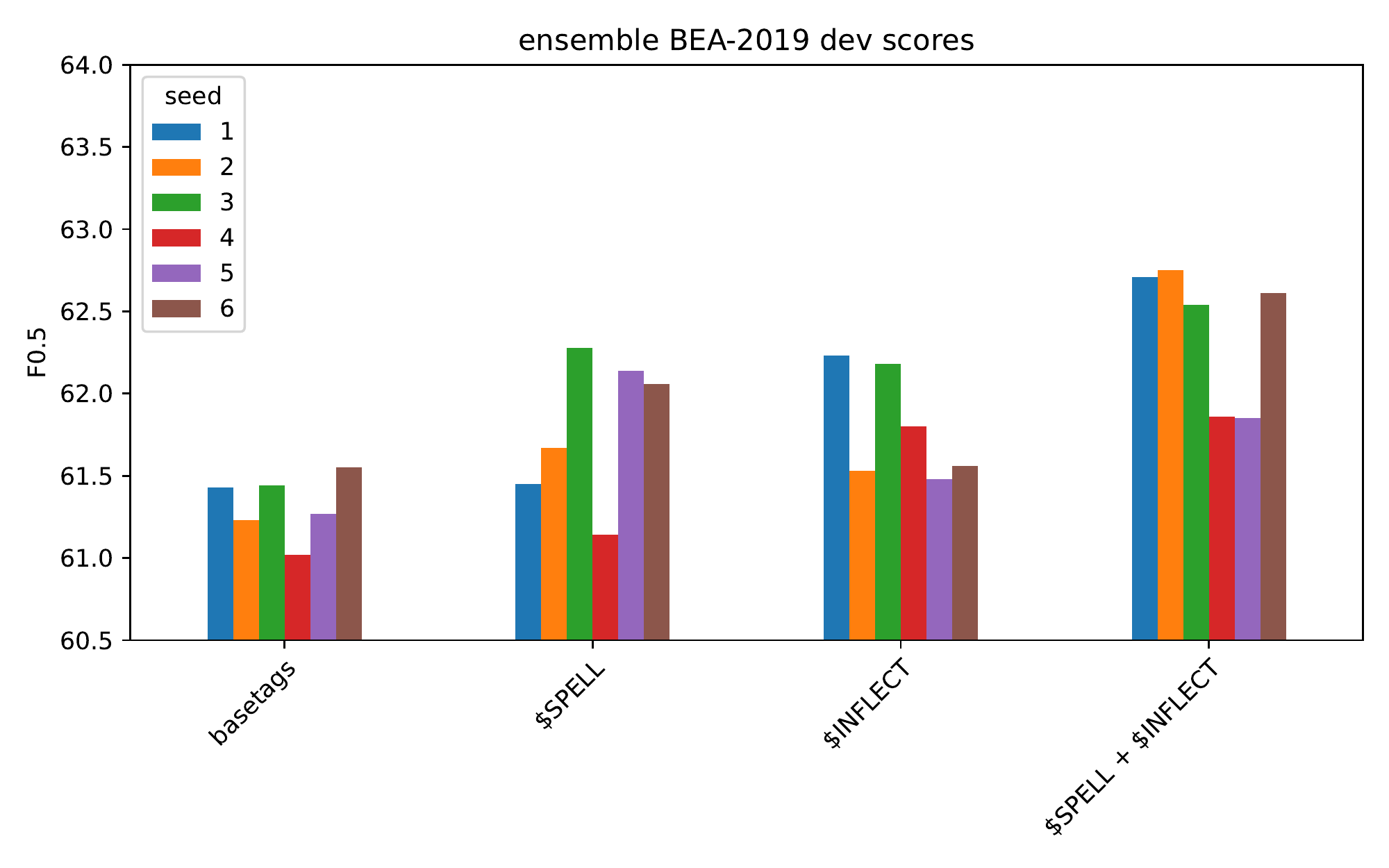}
    
    \caption{\label{fig:ensemble-dev-single} A bar graph showing the BEA-2019 development set $F_{0.5}$ scores of our ensemble models using different tagsets with six different random seeds. Each model is an ensemble of three encoders: $\text{DeBERTa}_{5K}^{(L)} \bigoplus \text{ELECTRA}_{5K}^{(L)} \bigoplus \text{RoBERTa}_{5K}^{(L)}$}
\end{figure}

\begin{figure}[t]
    \includegraphics[width=\linewidth]{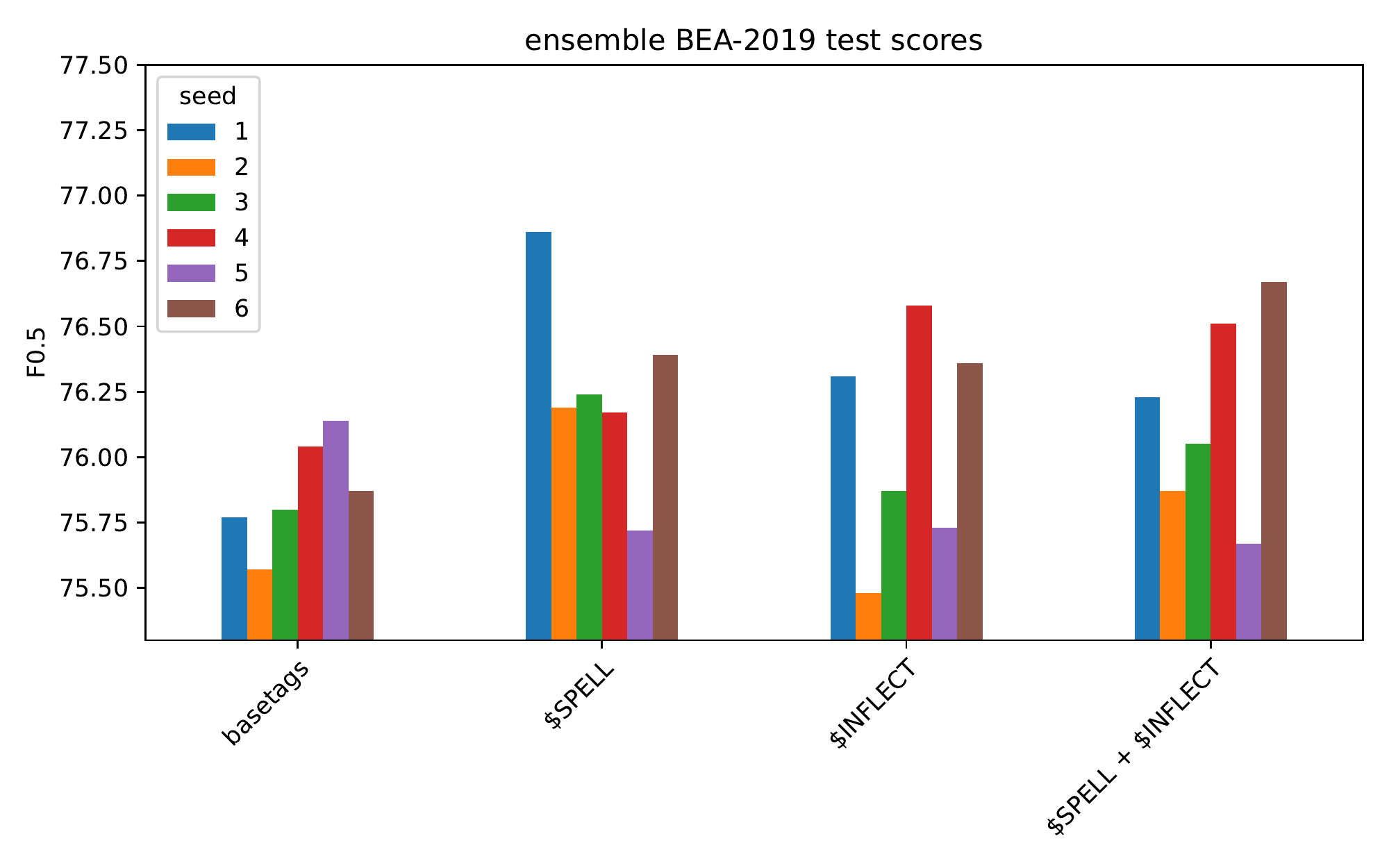}
    
    \caption{\label{fig:ensemble-test-single} A bar graph showing the BEA-2019 test set $F_{0.5}$ scores of our ensemble models using different tagsets with six different random seeds. Each model is an ensemble of three encoders: $\text{DeBERTa}_{5K}^{(L)} \bigoplus \text{ELECTRA}_{5K}^{(L)} \bigoplus \text{RoBERTa}_{5K}^{(L)}$}
\end{figure}

\begin{figure}[h!]
    \includegraphics[width=\linewidth]{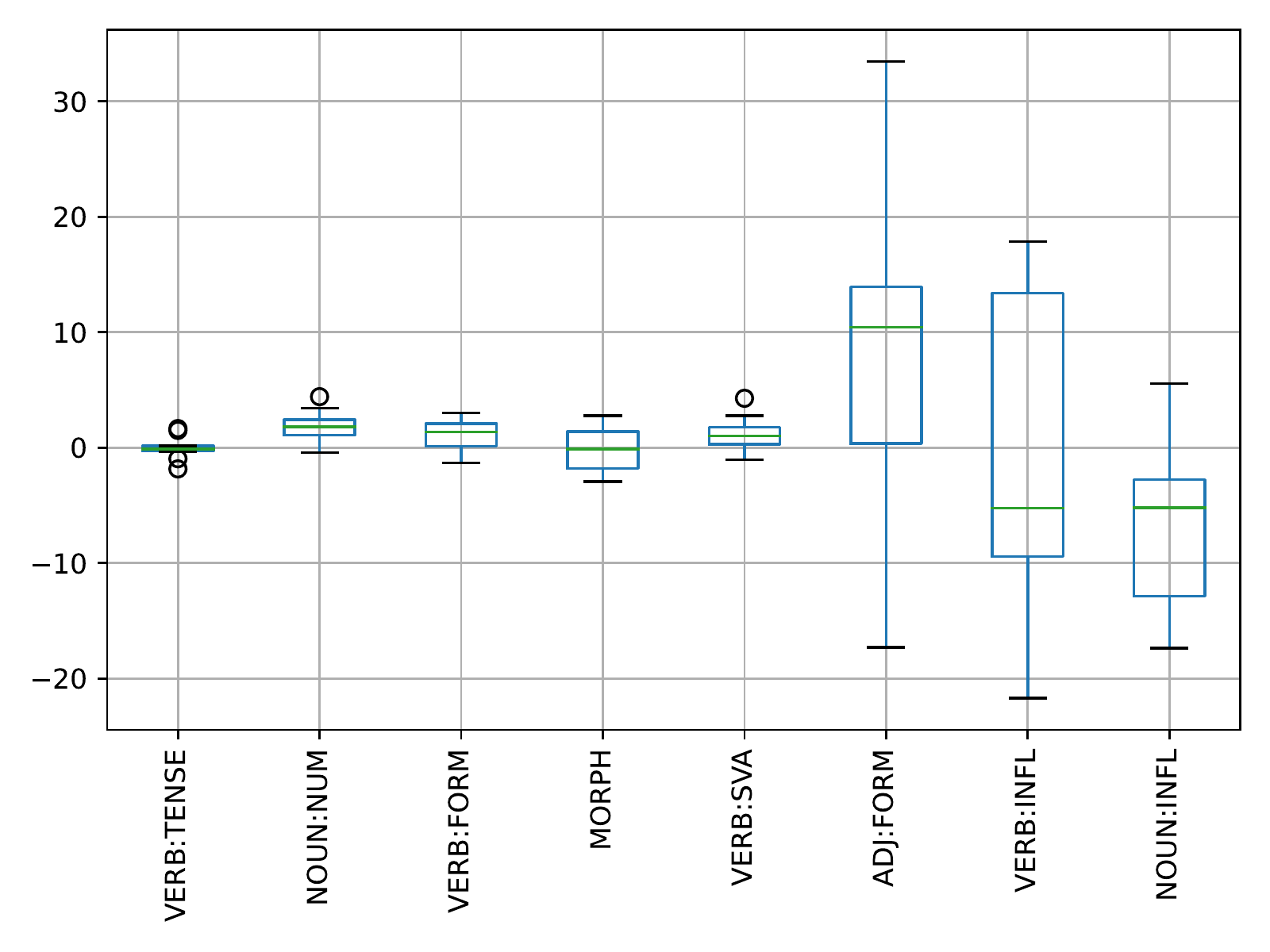}
    
    \cprotect\caption{\label{fig:lemon-per-category-target-error-scores} A box plot showing the change in BEA-2019 development set $F_{0.5}$ score for specific error categories when the \verb|$INFLECT| tagset is used instead of basetags. Each result shows the distribution of deltas over 10 combinations of encoders and tagset sizes. For each such combination and tagset, we take the mean $F_{0.5}$ score over three seeds and subtract the \verb|$INFLECT| mean from the basetags mean. The categories are ordered by frequency, decreasing from left to right.}
\end{figure}

\subsection{Additional Single Encoder and Ensemble Results}
\label{sec:additional-single-encoder-results}

For reference, we include the BEA-2019 development set scores of all of our single-encoder models in Figure~\ref{fig:single-model-dev-mean} and Table~\ref{tab:single-model-bea-dev-scores}. These models were trained as part of our search process for the best combination of encoders for our ensemble.

We also show, for individual seeds, the ensemble BEA-2019 development and test set scores in Figures \ref{fig:ensemble-dev-single} and \ref{fig:ensemble-test-single} respectively. This illustrates the variance in $F_{0.5}$ score over different random seeds.

% \usepackage{booktabs}
\begingroup
\renewcommand{\arraystretch}{1.5}

\begin{table*}[hp]
\centering
\begin{tabular}{@{}ccccc@{}}
\toprule
encoder \& tagset size & basetags     & \verb|$SPELL|      & \verb|$INFLECT|    & \verb|$SPELL| + \verb|$INFLECT| \\ \midrule
$\text{DeBERTa}_{10K}^{(L)}$    & 58.91 ± 0.62 & 59.16 ± 0.16 & 59.11 ± 0.52 & 60.00 ± 0.36         \\
$\text{DeBERTa}_{5K}^{(L)}$     & 59.02 ± 0.19 & \textbf{59.60 ± 0.56}  & 59.36 ± 0.49 & 60.04 ± 0.49        \\
$\text{DeBERTaV3}_{10K}^{(L)}$  & \textbf{59.25 ± 0.64} & 59.49 ± 0.39 & \textbf{59.41 ± 0.09} & 59.77 ± 0.19        \\
$\text{DeBERTaV3}_{5K}^{(L)}$   & 59.25 ± 0.53 & 58.56 ± 1.74 & \textbf{59.41 ± 0.07} & \textbf{60.10 ± 0.22}         \\
$\text{ELECTRA}_{10K}^{(L)}$    & 56.89 ± 0.15 & 57.32 ± 0.37 & 57.72 ± 0.12 & 58.17 ± 0.17        \\
$\text{ELECTRA}_{5K}^{(L)}$     & 56.83 ± 0.22 & 57.74 ± 0.35 & 57.71 ± 0.05 & 58.38 ± 0.31        \\
$\text{RoBERTa}_{10K}^{(L)}$    & 58.99 ± 0.38 & 59.11 ± 0.22 & 58.98 ± 0.59 & 59.47 ± 0.92        \\
$\text{RoBERTa}_{5K}^{(L)}$     & 58.63 ± 0.16 & 59.31 ± 0.22 & 59.26 ± 0.32 & 59.50 ± 0.77         \\
$\text{XLNet}_{10K}^{(L)}$      & 59.00 ± 0.36  & 58.65 ± 0.36 & 58.41 ± 0.40  & 58.69 ± 0.58        \\
$\text{XLNet}_{5K}^{(L)}$       & 58.20 ± 0.31  & 58.76 ± 0.19 & 58.29 ± 0.43 & 59.02 ± 0.20         \\ \bottomrule
\end{tabular}
\caption{A table showing BEA-2019 development set $F_{0.5}$ scores of single models using different tagsets and encoders. We show the mean and standard deviation of the scores over three training runs with different seeds.}
    \label{tab:single-model-bea-dev-scores}
\end{table*}
\endgroup
% \usepackage{booktabs}
\begin{table*}[hp]
    \centering
    \begin{tabular}{@{}p{19 em}|ccc@{}}
    \toprule
                                                                 & \multicolumn{3}{c}{CoNLL-2014 test}                                          \\
    Model                                                        & precision      & recall           & $F_{0.5} \text{ } (\bar{x} \pm \sigma)$  \\ \midrule
    $\text{DeBERTa}_{5K}^{(L)}$ basetags                         & 76.70          & 42.73            & 66.16 ± 0.47                             \\
    $\text{DeBERTa}_{5K}^{(L)}$ \verb|$SPELL|                    & \textbf{77.15} & \textbf{43.19}   & \textbf{66.64 ± 0.40}                    \\
    $\text{DeBERTa}_{5K}^{(L)}$ \verb|$INFLECT|                  & 76.43          & 42.57            & 65.90 ± 0.49                             \\
    $\text{DeBERTa}_{5K}^{(L)}$ \verb|$SPELL| + \verb|$INFLECT|  & 76.62          & 42.67            & 66.06 ± 0.44                             \\ \midrule
    ensemble basetags                                            & 80.70          & 41.25            & 67.72 ± 0.32                             \\
    ensemble \verb|$SPELL|                                       & \textbf{80.86} & \textbf{41.72}   & \textbf{68.06 ± 0.43}                    \\
    ensemble \verb|$INFLECT|                                     & 80.60          & 41.31            & 67.70 ± 0.54                             \\
    ensemble \verb|$SPELL| + \verb|$INFLECT|                     & 80.65          & 41.70            & 67.93 ± 0.40                             \\ \midrule
    $\text{DeBERTa}_{10K}^{(L)} \bigoplus \text{RoBERTa}_{10K}^{(L)} \bigoplus \text{XLNet}_{5K}^{(L)}$ \citep{tarnavskyi-etal-2022-ensembling} & 76.1                     & 41.6                     & 65.3                              \\
    $\text{RoBERTa}_{5K}^{(L)}$ (KD)  \citep{tarnavskyi-etal-2022-ensembling}                      & 74.40                        & 41.05                       & 64.0                               \\
    T5 xxl \citep{rothe-etal-2021-simple}           & -                         & -                         & 68.87                             \\
    ESC \citep{qorib-etal-2022-frustratingly}       & \textbf{81.48}            & \textbf{43.78}            & \textbf{69.51}               \\ \bottomrule
    \end{tabular}
    \caption{A table showing CoNLL-2014 test set scores (using the $\text{M}^2$ scorer). The top section shows our models with varying tagsets using the $\text{DeBERTa}_{5K}^{(L)}$ encoder. The middle section shows the results for our ensemble models with varying tagsets. In the table, "ensemble" denotes the encoders $\text{DeBERTa}_{5K}^{(L)} \bigoplus \text{ELECTRA}_{5K}^{(L)} \bigoplus \text{RoBERTa}_{5K}^{(L)}$. Finally, the bottom section shows models from related work. The model labelled "(KD)" was trained using \citet{tarnavskyi-etal-2022-ensembling}'s knowledge distillation procedure. The results in the top and middle sections are averaged over 6 seeds, and the standard deviation, $\sigma$, of the test $F_{0.5}$ is shown.}
    \label{tab:combined-conll-results}
    \end{table*}

\subsection{Performance Analysis of Inflection-Related Error Categories}
\label{sec:detailed-inflection-error-analysis}

To illustrate which of its target error categories the \verb|$INFLECT| tagset has successfully improved on, Figure~\ref{fig:lemon-per-category-target-error-scores} shows, for each error category, the distributions of the difference in BEA-2019 development set scores between models using the \verb|$INFLECT| and basetags tagsets over all 10 models (5 encoders, each with vocab sizes of 5k and 10k). We observe that the \verb|ADJ:FORM|, \verb|VERB:INFL| and \verb|NOUN:INFL| have a very high range of differences. This is expected because these three categories have frequencies of 11, 6 and 4 respectively in the development set. The small sample size makes it difficult to draw conclusions about these error categories. By contrast, the remaining five categories shown in the Appendix in Figure~\ref{fig:lemon-per-category-target-error-scores} have development set frequencies ranging from 478 for \verb|VERB:TENSE| to 141 for \verb|VERB:SVA|. Within these high-frequency categories, we observe that the \verb|NOUN:NUM|, \verb|VERB:FORM| and \verb|VERB:SVA| have positive median changes.

\subsection{CoNLL-2014 Results}
\label{sec:conll-2014-results}

For interested readers, we have included results on the CoNLL-2014 benchmark \citep{ng-etal-2014-conll} in Table~\ref{tab:combined-conll-results}. The scores are computed with the $\text{M}^2$ scorer \citep{dahlmeier-ng-2012-better}. In both the single and ensemble models, the \verb|$SPELL| tagset performs best. However, these results should be interpreted with caution, since the model hyper-parameters were not tuned on the CoNLL-2014 development set.

\end{document}